\documentclass[10pt,twocolumn,letterpaper]{article}

\usepackage{iccv}
\usepackage{times}
\usepackage{epsfig}
\usepackage{graphicx}
\usepackage{amsmath}
\usepackage{amssymb}

\usepackage{url}            %
\usepackage{booktabs}       %
\usepackage{amsfonts}       %
\usepackage{nicefrac}       %
\usepackage{microtype}      %
\usepackage{xcolor}         %

\usepackage{pifont}
\usepackage{color}
\usepackage{wrapfig}
\usepackage{makecell}
\usepackage{colortbl}
\usepackage{multirow}
\usepackage{fixltx2e}
\usepackage{subcaption}
\usepackage{etoolbox}
\usepackage{multicol}
\usepackage{refcount}
\usepackage{enumitem}
\usepackage{pifont}
\usepackage[misc]{ifsym}
\usepackage[accsupp]{axessibility}

\newcommand{\cmark}{\ding{51}}%
\newcommand{\xmark}{\ding{55}}%
\definecolor{LightCyan}{rgb}{0.88,1,1}
\definecolor{mygray}{gray}{0.9}
\definecolor{mygray2}{gray}{0.6}

\usepackage[pagebackref=true,breaklinks=true,letterpaper=true,colorlinks,bookmarks=false]{hyperref}

\iccvfinalcopy %

\def\iccvPaperID{11844} %

\iccvfinaltrue

\ificcvfinal\fi

\hypersetup{
    colorlinks=true,
    linkcolor=blue,
    filecolor=magenta,      
    urlcolor=cyan,
    pdftitle={MixMAE},
    pdfpagemode=FullScreen,
    }

\begin{document}

\title{GeoMIM: Towards Better 3D Knowledge Transfer via Masked Image Modeling for Multi-view 3D Understanding}
\author{%
  Jihao Liu$^{1,2}$ \quad Tai Wang$^{1,3}$ \quad Boxiao Liu$^{2}$ \quad Qihang Zhang$^{1}$ \quad Yu Liu$^{2}$~\textsuperscript{\Letter} \quad Hongsheng Li$^{1,3,4}$~\textsuperscript{\Letter} \\
  $^1$CUHK MMLab \quad
  $^2$SenseTime Research  \\
  $^3$Shanghai AI Laboratory \quad
  $^4$CPII under InnoHK
}

\maketitle
\ificcvfinal\thispagestyle{empty}\fi

\makeatletter{\renewcommand*{\@makefnmark}{}
\footnotetext{\textsuperscript{\Letter} Corresponding author.}\makeatother}

\begin{abstract}
    Multi-view camera-based 3D detection is a challenging problem in computer vision. Recent works leverage a pretrained LiDAR detection model to transfer knowledge to a camera-based student network. However, we argue that there is a major domain gap between the LiDAR BEV features and the camera-based BEV features, as they have different characteristics and are derived from different sources. In this paper, we propose Geometry Enhanced Masked Image Modeling (GeoMIM) to transfer the knowledge of the LiDAR model in a pretrain-finetune paradigm for improving the multi-view camera-based 3D detection. GeoMIM is a multi-camera vision transformer with Cross-View Attention (CVA) blocks that uses LiDAR BEV features encoded by the pretrained BEV model as learning targets. During pretraining, GeoMIM's decoder has a semantic branch completing dense perspective-view features and the other geometry branch reconstructing dense perspective-view depth maps. 
    The depth branch is designed to be camera-aware by inputting the camera's parameters for better transfer capability. Extensive results demonstrate that GeoMIM outperforms existing methods on nuScenes benchmark, achieving state-of-the-art performance for camera-based 3D object detection and 3D segmentation. Code and pretrained models are available at \url{https://github.com/Sense-X/GeoMIM}.
\end{abstract}

\section{Introduction}
\label{sec:intro}

Multi-view camera-based 3D detection is an emerging critical problem in computer vision~\cite{bevdet,fcos3d,detr3d,bevdet4d,li2022bevdepth,bevformer,solofusion,bevstereo,petr,petrv2,shao2023safety,shao2023reasonnet}. To improve the detection performance, recent works~\cite{monodistill,uvtr,tigbev} often choose to use a pretrained LiDAR model as the teacher and transfer its knowledge to a camera-based student network. Various techniques, such as LIGA-Stereo~\cite{liga}, CMKD~\cite{cmkd}, and BEVDistill~\cite{bevdistill}, have been proposed to leverage the rich geometry information of the LiDAR model's BEV (bird's eye view) features.

Utilizing a pretrained LiDAR model to provide auxiliary supervision has become a widely adopted design that can enhance the performance of camera-based models. However, we contend that this design is not optimal due to a significant domain gap between the BEV features of the LiDAR model and those of the camera-based model. This domain gap arises from the 3D and sparse characteristics of LiDAR point clouds compared to the dense 2D images captured by the camera. Additionally, the LiDAR model's BEV features are grounded in ground truth depth, while those of the camera-based model are typically inferred from 2D images, a problem that is often ill-posed.
We empirically demonstrate their domain gap with a pilot study as shown in Tab.~\ref{tab:pivot}. We find that utilizing a LiDAR teacher to provide auxiliary supervision can indeed improve an ImageNet-pretrained~\cite{swin} camera-based model, but is unable to improve a stronger camera-based model initialized by recent powerful self-supervised pretraining. In other words, directly utilizing the pretrained LiDAR model to distill the final camera-based model might not be an optimal design and does not necessarily lead to performance gain.

\begin{table}[t]
    \centering
    \resizebox{0.9\linewidth}{!}{
    \begin{tabular}{ll|cc} 
    \toprule
    \textbf{Pretrain} & \textbf{Supervision} & \textbf{Finetune} & \textbf{\makecell{Finetune + \\ LiDAR BEV}} \\
    \midrule
    SL~\cite{swin} & Classes & 40.6 & 41.7  \\
    SSL~\cite{mixmim} & RGB Pixels &  44.3 & 43.9 \\
    GeoMIM & BEV Feature & \textbf{47.2} & 45.4 \\
    \bottomrule
    \end{tabular}}
    \caption{The effects of LiDAR BEV feature distillation on ImageNet-pretrained (SL), self-supervised (SSL), and our GeoMIM pretraining-finetuning settings for BEVDet in nuScenes 3D detection. Naively distilling LiDAR BEV features in finetuning introduces domain gaps and harms the performance when the pretrianed model is powerful enough.}
    \vspace{-1em}
    \label{tab:pivot}
\end{table}

To better take advantage of the LiDAR model, in this paper, we propose \emph{Geometry Enhanced Masked Image Modeling (GeoMIM)} to transfer the knowledge of the LiDAR model in a pretrain-finetune paradigm for improving the multi-view camera-based 3D detection. It is built upon a multi-camera vision transformer with \emph{Cross-View Attention (CVA)} blocks and enables perspective-view (PV) representation pretraining via BEV feature reconstruction from masked images. Specifically, during pretraining, we partition the training images into patches and feed a portion of them into the encoder following Masked Autoencoder~\cite{mae}. Our GeoMIM decoder then uses these encoded visible tokens to reconstruct the pretrained LiDAR model's BEV feature in the BEV space instead of commonly used RGB pixels~\cite{simmim,mae,mixmim} or depth points~\cite{multimae} as in existing MAE frameworks. To achieve this PV to BEV reconstruction, we first devise two branches to \emph{decouple} the semantic and geometric parts, with one branch completing dense PV features and the other reconstructing the depth map. The dense PV features can then be projected into the BEV space with the depth distribution following Lift-Splat-Shoot (LSS)~\cite{lss}. We further equip the two branches with the proposed CVA blocks in their intermediate layers to allow each patch to attend to tokens in other views. It enhances the decoder's capability of joint multi-view inference which is especially critical for BEV feature reconstruction. Finally, the depth branch is designed to be \emph{camera-aware} with the additional encoding of cameras' parameters as input, making the pretrained GeoMIM better adapt to downstream tasks with different cameras.

To demonstrate the effectiveness of GeoMIM, we finetune the pretrained backbone to conduct multi-view camera-based 3D detection and 3D segmentation on the nuScenes~\cite{nuscenes} dataset. We achieve state-of-the-art results of 64.4 NDS (NuScenes Detection Score) and 70.5 mIoU (mean intersection over union) for 3D detection and segmentation on the NuScenes \texttt{test} set, which are 2.5\% and 1.1\% better than previously reported best results~\cite{solofusion,tpvformer}. Additionally, we verify that the backbone pretrained on nuScenes dataset can be successfully transferred to Waymo Open dataset~\cite{waymo}, improving the mAP (mean average precision) of the ImageNet-initialized 3D detector by 6.9\%.

\begin{figure*}
    \centering
    \includegraphics[width=1.0\linewidth]{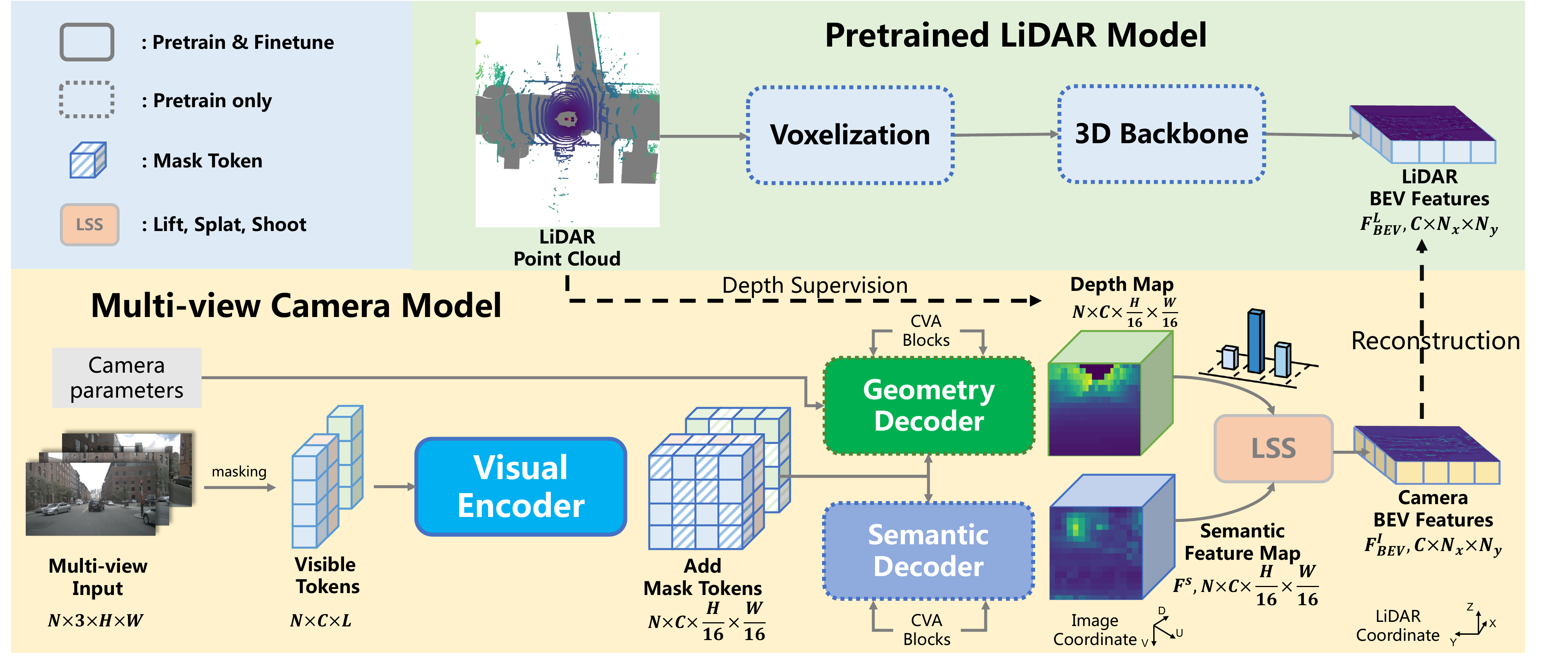}
    \vspace{-2em}
    \caption{Overview of GeoMIM. For pretraining, the multi-view images are randomly masked for a proportion of image tokens, and only the visible tokens are processed by the encoder. Right before decoding, the token embeddings are filled with mask tokens for separately decoding dense camera-view semantic features and depth maps, which are then projected to BEV space for reconstructing the LiDAR BEV features. After pretraining, only the encoder is finetuned on downstream tasks.}
    \label{fig:framework}
\end{figure*}

\begin{figure}
    \centering
    \includegraphics[width=1.0\linewidth]{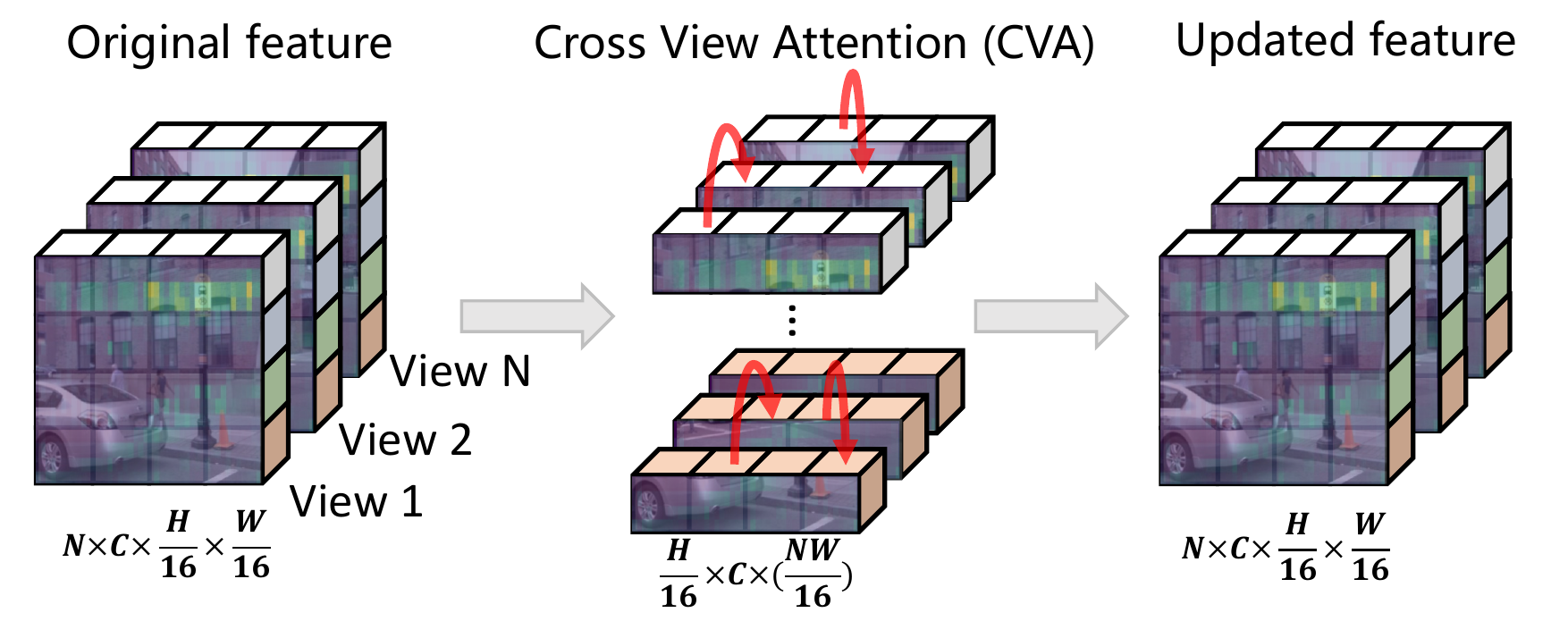}
    \vspace{-2em}
    \caption{Cross-view attention block. We partition the multi-view inputs into multiple groups according to their row indices, and perform self-attention within each group.}
    \vspace{-1em}
    \label{fig:cva}
\end{figure}

\section{Related Works}

\noindent\textbf{Masked Image Modeling}\quad Inspired by BERT~\cite{bert} for Masked Language Modeling, Masked Image Modeling (MIM) becomes a popular pretext task for visual representation learning~\cite{beit,mae,sit,MaskedVP,msn,data2vec,ibot,multimae,zhang2022point}. MIM aims to reconstruct the masked tokens from a corrupted input. SimMIM~\cite{simmim} points out that raw pixel values of the randomly masked patches are a good reconstruction target and a lightweight prediction head is sufficient for pretraining. Different from SimMIM, MAE~\cite{mae} only takes the visible patches as the input of the encoder. Mask tokens are added in the middle of the encoder and the decoder. BEiT~\cite{beit} utilizes a pretrained discrete VAE (dVAE)~\cite{dvae,dalle} as the tokenizer. PeCo~\cite{peco} proposed to apply perceptual similarity loss on the training of dVAE can drive the tokenizer to generate better semantic visual tokens, which helps pretraining. In contrast to those works, our GeoMIM utilizes a geometry-rich LiDAR model and transfers its knowledge via MIM pretraining, aiming to improve the multi-view camera-based 3D models.

\noindent\textbf{Multi-view camera-based 3D detection}\quad The field of camera-based 3D object detection has seen significant progress in recent years~\cite{fcos3d,detr3d,bevformer,bevdet,li2022bevdepth,bevstereo,solofusion,zhang2022monodetr}. FCOS3D~\cite{fcos3d} proposed a fully convolutional single-stage detector for monocular 3D object detection. DETR3D~\cite{detr3d} extends the DETR framework to the 3D domain, and proposes a framework for end-to-end 3D object detection.
BEVFormer~\cite{bevformer} combines BEV (bird’s eye view) representation and transformer networks for 3D object detection.
BEVDepth~\cite{li2022bevdepth} focuses on accurately estimating the depth of objects in the BEV representation.
Additionally, considering the promising performance of the LiDAR-based detectors, there are several papers that use a pretrained LiDAR detector for knowledge distillation~\cite{kd}. 
LIGA-Stereo~\cite{liga} proposes to mimic the LiDAR BEV features for training a camera-based detector. UVTR~\cite{uvtr} represents different modalities in a unified manner and supports knowledge transfer with the voxel representations. More recent BEVDistill~\cite{bevdistill} and CMKD~\cite{cmkd} not only use the LiDAR BEV features for knowledge distillation but also transfer the teacher's knowledge through sparse instance distillation and response-based distillation respectively. 
In comparison, we utilized the pretrained LiDAR model in a pretraining-finetuning paradigm to avoid the LiDAR-camera BEV domain gap.

\section{Method}

Employing a pretrained LiDAR-based detection model to provide auxiliary learning guidance to train camera-based 3D understanding models has shown promising results in recent years~\cite{liga,bevdistill,monodistill,uvtr,cmkd}. However, because of the domain gap between the LiDAR and camera modalities, we observe that when a camera-based model is already strong, directly supervising it with the LiDAR teacher fails to improve the camera-based model as shown in Tab.~\ref{tab:pivot}. 

To address this problem, we propose GeoMIM to better transfer the LiDAR model's knowledge to the camera-based model in a pretrain-finetune paradigm. 
GeoMIM pretrains a multi-view camera-based model via Masked Image Modeling (MIM)~\cite{simmim}. 
Unlike existing 2D MAE works~\cite{mae,mixmim}, we project the semantic features to the BEV (bird's eye view) space and use the LiDAR BEV features in the 3D space as the reconstruction targets for pretraining. 
The pretrained LiDAR model is only used in the pretraining stage, and is discarded in the finetuning stage to avoid introducing the LiDAR-camera BEV domain gap. We illustrate the proposed GeoMIM in Fig.~\ref{fig:framework}.

\noindent\textbf{Masking and Encoder}\quad
Given the multi-view input images $X = \{x_i \in \mathbb{R}^{3 \times H \times W }, i=1,2,\dots,N\}$ where $N$, $H$, $W$ are the number of views, image height, and width, we randomly mask a proportion of input image patches (tokens) and use a Swin Transformer~\cite{swin} as the encoder to encode the visible tokens. The encoded representations, $F^v \in \mathbb{R}^{N \times C \times L}$ where $C$ and $L$ denote the number of dimensions and the number of visible tokens, are then filled with a shared mask token $\mathrm{[M]} \in \mathbb{R}^{C}$ at the masked locations and further processed by the decoder for reconstruction.

\noindent\textbf{GeoMIM Decoder}\quad
To transfer the rich geometry knowledge of a pretrained LiDAR detector to our camera-based model, we jointly project the multi-view semantic features according to their estimated depth maps to the BEV space and use the same scene's LiDAR BEV features as the reconstruction targets. Specifically, our GeoMIM uses two \emph{decoupled} decoders, each of which consists of 8 Transformer~\cite{attention} blocks. 
The semantic decoder $\mathrm{D_{sem}}$ reconstructs the dense camera-view semantic features $F^s \in \mathbb{R}^{N \times C \times \frac{H}{16} \times \frac{W}{16}}$ of the $N$ camera views and the other geometry decoder $\mathrm{D_{geo}}$ predicts dense camera-view depth maps $D \in \mathbb{R}^{N \times B \times \frac{H}{16} \times \frac{W}{16}}$ of the $N$ camera views, where $B$ denotes the number of depth bins. 
The depth map and semantic feature can be expressed as
\begin{equation}
\label{eq:dec}
    D = \mathrm{D_{geo}}(F^v, \mathrm{[M]}), \quad F^s = \mathrm{D_{sem}}(F^v, \mathrm{[M]}).
\end{equation}
We can then obtain the camera BEV features $F_{BEV}^I$ by jointly projecting the multi-view semantic features to the BEV space with the Lift-Splat-Shoot (LSS)~\cite{lss} operation according to the predicted dense depth maps,
\begin{equation}
\label{eq:lss}
    F_{BEV}^I \in \mathbb{R}^{C \times N_x \times N_y} = \mathrm{LSS}(F^s, D),
\end{equation}
where $N_x$, $N_y$ are the numbers of bins in the $x$ and $y$ axis of the BEV feature maps respectively. 
Empirically, the two decoders share the first half of the Transformer blocks for efficiency. 

Unlike existing works that separately process the multi-view input images,  we propose a novel \emph{Cross-View Attention (CVA)} block to model the interaction across different views to better reconstruct the LiDAR BEV features from input images. Our intuition is that as the multi-view images are naturally overlapped, proper interaction across views is beneficial to align those images and better infer the LiDAR BEV features.
Instead of explicitly using the epipolar lines to associate pixels across the multi-view images,
we partition the camera-view tokens of the multiple views into groups according to their row indices and only allow the tokens belonging to the same row of the $\frac{1}{16}$ input resolution to interact with each other. 
The interaction is modeled by the self-attention operation~\cite{attention}. Notably, our proposed CVA has linear computation complexity to the input image size and is therefore much more efficient compared to global self-attention. We illustrate the proposed CVA in Fig.~\ref{fig:cva}. 
We use the CVA block as the $2$th and $6$th attentions blocks of the decoder. Note that we do not add it to the backbone and no extra computation is introduced when finetuning the encoder.

Accurately reconstructing depth with the geometry decoder implicitly requires the decoder to infer the camera's intrinsic parameters, which is difficult to generalize to an unseen dataset as the data may be collected with different cameras. To achieve better transferability across different downstream tasks, 
we encode the camera's intrinsic and extrinsic parameters using a linear projection layer and use the resulting features to scale the geometry decoder's feature using the Squeeze-and-Excitation module~\cite{se}. Importantly, we do not require the camera's information when finetuning on downstream tasks since only the decoder uses the camera's information during pretraining. We demonstrate that the camera-aware depth reconstruction branch leads to better performance when finetuning on tasks that differ from the pretraining dataset.

\noindent\textbf{Loss}\quad
We use the mean squared error (MSE) loss between the projected camera BEV features and the pretrained LiDAR BEV features for pretraining,
\begin{equation}
\label{eq:loss_rec}
    \mathcal{L}_\mathnormal{rec} = \Vert(F_{BEV}^I - F_{BEV}^L) \Vert_2^2,
\end{equation}
where $F_{BEV}^L \in \mathbb{R}^{C \times N_x \times N_y}$ denotes the pretrained LiDAR model's BEV features. 
In addition, we incorporate a depth prediction task.
Following prior arts~\cite{li2022bevdepth}, we use the ground truth discrete depth $D_{GT}$ derived from the LiDAR point cloud and calculate the binary cross entropy (BCE) loss as the depth loss,
\begin{equation}
\label{eq:loss_depth}
    \mathcal{L}_\mathnormal{depth} = \mathrm{BCE}(D, D_{GT}).
\end{equation}
The overall loss can be expressed as
\begin{equation}
\label{eq:loss}
    \mathcal{L} = \mathcal{L}_\mathnormal{rec} + \alpha\mathcal{L}_\mathnormal{depth},
\end{equation}
where $\alpha$ balances the two loss terms, which is set as 0.01 experimentally. Empirically, we observe that the depth loss can enhance the convergence speed, which is crucial for pretraining large models. 

After pretraining, we discard the decoders and add a task-specific head on the top of the encoder for downstream tasks finetuning. During finetuning, we only utilize ground-truth supervision and abstain from utilizing the LiDAR model to avoid introducing the aforementioned domain gap.

\noindent\textbf{Comparison with 2D MAE}\quad
Compared to existing 2D MAE models~\cite{mae,simmim,mixmim}, our proposed GeoMIM's pretraining has two distinct characteristics: (1) We employ a geometry-rich LiDAR model and transfer its high-level knowledge in the BEV space via MIM pretraining, which can effectively enhance the geometry perception capability of the camera-based model. 
In contrast, the original MAE~\cite{mae} reconstructs image pixels and could work well for 2D downstream perception tasks, but is found to be less effective for 3D perception. The reason is that the autonomous driving dataset, e.g., nuScenes~\cite{nuscenes}, is much less diverse than MAE's pretraining dataset ImageNet-1K~\cite{imagenet}. As a result, employing image pixel reconstruction as the pretext task is hard to learn high-quality representations.
(2) Contrary to MAE which only calculates the reconstruction loss in the masked tokens, we take all tokens into consideration in our loss. This is because the learning targets we use are from a different modality and in a different geometric space. We can take full advantage of the LiDAR model by using all tokens to calculate the loss. For the masked locations, the objective is a prediction task while for the unmasked locations, it is similar to a distillation task.

\begin{table*}
    \centering
    \resizebox{\textwidth}{!}{
    \begin{tabular}{l|c|c|c|c|c|c|c@{\hspace{1.0\tabcolsep}}c@{\hspace{1.0\tabcolsep}}c@{\hspace{1.0\tabcolsep}}c@{\hspace{1.0\tabcolsep}}c} 
    
    \toprule
    \textbf{Framework} & \textbf{Pretrain} & \textbf{Backbone} & \textbf{Image Size} & \textbf{CBGS} & \textbf{mAP}$\uparrow$  &\textbf{NDS}$\uparrow$  & \textbf{mATE}$\downarrow$ & \textbf{mASE}$\downarrow$   &\textbf{mAOE}$\downarrow$   &\textbf{mAVE}$\downarrow$   &\textbf{mAAE}$\downarrow$  \\
    \midrule
    
    DETR3D~\cite{detr3d} & \multirow{4}{*}{FCOS3D} & \multirow{4}{*}{R101-DCN} & 900 $\times$ 1600 & \cmark & 0.349 & 0.434 & 0.716 & 0.268 & 0.379 & 0.842 & 0.200 \\
    
    BEVFormer~\cite{bevformer} &  &  & 900 $\times$ 1600 & \xmark & 0.416 & 0.517 & 0.673 & 0.274 & 0.372 & 0.394 & 0.198 \\

    UVTR~\cite{uvtr} &  &  & 900 $\times$ 1600 & \xmark & 0.379 & 0.483 & 0.731 & 0.267 & 0.350 & 0.510 & 0.200 \\
    
    PolarFormer~\cite{polarformer} &  &  & 900 $\times$ 1600 & \xmark & 0.432 & 0.528 & 0.648 & 0.270 & 0.348 & 0.409 & 0.201 \\
    
    \midrule
    
    PETR~\cite{petr} & \multirow{3}{*}{ImageNet} & \multirow{3}{*}{R101} & 512 $\times$ 1408 & \cmark & 0.357 & 0.421 & 0.710 & 0.270 & 0.490 & 0.885 & 0.224 \\
    
    PETRv2~\cite{petrv2} &  &  & 640 $\times$ 1600 & \cmark & 0.421 & 0.524  & 0.681 & 0.267 & 0.357 & 0.377 & 0.186 \\

    SOLOFusion~\cite{solofusion} &  &  & 512 $\times$ 1408                          & \cmark & 0.483 & 0.582 & 0.503 & 0.264 & 0.381 & \textbf{0.246} & 0.207 \\
    
    \midrule
    \midrule
    
    BEVDepth~\cite{li2022bevdepth} & \multirow{2}{*}{ImageNet} & \multirow{2}{*}{ConvNeXt-B} & 512 $\times$ 1408 & \cmark & 0.462 & 0.558 & - & - & - & - & - \\ 
    
    BEVStereo~\cite{bevstereo} &  &  & 512 $\times$ 1408 & \cmark & 0.478 & 0.575 & - & - & - & - & - \\ 
    
    \midrule
    
    BEVDet4D~\cite{bevdet4d} & \multirow{2}{*}{ImageNet} & \multirow{2}{*}{Swin-B} & 640 $\times$ 1600 & \cmark & 0.421 & 0.545 & 0.579 &  \textbf{0.258} &  \textbf{0.329} &  0.301 &  \textbf{0.191} \\
    BEVDepth\textsuperscript{$\dagger$} &  &  & 512 $\times$ 1408 & \cmark & 0.466 & 0.555 & 0.531 &  0.264 &  0.489 &  0.293 &  0.200 \\
    
    \midrule
    \rowcolor[gray]{.9} 
    BEVDepth & GeoMIM & Swin-B & 512 $\times$ 1408                          & \cmark & \textbf{0.523} & \textbf{0.605} & \textbf{0.470} & 0.260 & 0.377 & 0.254 & 0.195 \\ 
    \bottomrule
    \end{tabular}}
    
    \vspace{-0.8em}
    \caption{Comparison on nuScenes \texttt{val} set. $\dagger$ denotes our implementation with the official code.}
    \label{tab:main_val_set}
\end{table*}

\begin{table*}[t]
    \centering
    \resizebox{0.73\linewidth}{!}{
    \begin{tabular}{l|cc|cc|cc} 
    \toprule
    \multirow{2}{*}{\textbf{Pretrain}} & \multicolumn{2}{c|}{\textbf{3D-Segmentation}} & \multicolumn{2}{c|}{\textbf{Waymo}} & \multicolumn{2}{c}{\textbf{nuImages}} \\
     & mIoU\textsuperscript{val} & mIoU\textsuperscript{test} & LET-3D APL & LET-3D AP & AP\textsuperscript{box} & AP\textsuperscript{mask} \\

    \midrule
    & \multicolumn{2}{c|}{TPVFormer~\cite{tpvformer}} & \multicolumn{2}{c|}{DfM~\cite{dfm}} & \multicolumn{2}{c}{Mask-RCNN~\cite{maskrcnn}} \\
    \midrule
    Supervised~\cite{swin} & 66.4 & 68.3 & 31.5 & 44.6 & 49.0 & 41.3 \\
    Self-supervised~\cite{mixmim} & 65.0 & 66.3 & 34.8 & 49.5 & 51.5 & 41.8 \\
    \rowcolor[gray]{.9} 
    GeoMIM & \textbf{68.9} & \textbf{70.5} & \textbf{37.8} & \textbf{52.5} & \textbf{52.9} & \textbf{44.4} \\
    \bottomrule
    \end{tabular}}
    
    \vspace{-1em}
    \caption{Transfer learning results on 3D-segmentation with TPVFormer (left), Open Waymo 3D detection with DfM (middle), and nuImages object detection and segmentation with Mask-RCNN (right).}
    \label{tab:transfer}
\end{table*}

\begin{table}[t]
    \centering
    \resizebox{0.9\linewidth}{!}{
    \begin{tabular}{l|c|c|cc} 
    \toprule
    \textbf{Pretrain} & \textbf{NDS}$\uparrow$ & \textbf{mAP}$\uparrow$ & \textbf{mATE}$\downarrow$ & \textbf{mAOE}$\downarrow$ \\
    \midrule
    Supervised~\cite{swin} & 0.406 & 0.326 & 0.665 & 0.546 \\
    EsViT~\cite{esvit} & 0.389 & 0.305 & 0.699 & 0.516 \\
    UniCL~\cite{unicl} & 0.396 & 0.314 & 0.694 & 0.596 \\
    MixMAE~\cite{mixmim} & 0.443 & 0.374 & 0.647 & 0.418 \\
    \rowcolor[gray]{.9} 
    GeoMIM & \textbf{0.472} & \textbf{0.397} & \textbf{0.614} & \textbf{0.395} \\
    \bottomrule
    \end{tabular}}
    \vspace{-1em}
    \caption{Comparison with previous pretraining methods on nuScenes with BEVDet.}
    \vspace{-1.5em}
    \label{tab:pretrain}
\end{table}

\section{Experiment Setups}

To demonstrate the effectiveness of GeoMIM, we conduct experiments by pretraining Swin Transformer~\cite{swin} backbones with GeoMIM and then finetuning it on various downstream tasks. These tasks include multi-view camera-based 3D detection on nuScenes~\cite{nuscenes} and Open Waymo~\cite{waymo} datasets, camera-based 3D semantic segmentation on nuScenes dataset, and 2D detection on nuImages dataset.

\noindent\textbf{Dataset and Evaluation Metrics}\quad
We use the large-scale nuScenes dataset for pretraining and finetuning, which contains 750, 150, and 150 scenes for training, validation, and testing, respectively. Each scene has 6 camera images and LiDAR point cloud covering 360\textsuperscript{$\circ$}. Following the official evaluation metrics, we primarily report NuScenes Detection Score (NDS) and mean Average Precision (mAP) for comparison. We also report other five metrics, including ATE, ASE, AOE, AVE, and AAE, to measure translation, scale, orientation, velocity, and attribute errors, respectively, for a more detailed diagnosis.

We also evaluate the transferability of GeoMIM by finetuning the pretrained backbone on the Open Waymo and nuImages datasets. 
We report LET-3D-APL~\cite{hung2022let} and LET-3D-AP following the latest official guidelines for comparison. 
We report Mean Average Precision (mAP) of box and mask on nuImages dataset for 2D object detection and instance segmentation.

\noindent\textbf{Pretraining}\quad
We pretrain the Swin Transformer backbones on the training split of the nuScenes dataset with multi-view images as input. By default, we pretrain for 50 epochs with an input size of $256\times704$. For ablation studies, we pretrain for 6 epochs unless otherwise specified. We use a pretrained TransFusion-L~\cite{transfusion} LiDAR model to provide the reconstruction targets. We randomly mask the multi-view input images with a mask ratio of 50\%. We use AdamW~\cite{adamw} optimizer with a learning rate of $2 \times 10^{-4}$ and weight decay of 0.01. The learning rate is linearly warmed-up for 500 iterations and cosine decayed to 0. We apply the data augmentation strategy in BEVDet~\cite{bevdet} to augment the input images and do not use augmentations for the LiDAR inputs. We utilize the Swin-Base and -Large backbones for pretraining, initializing the backbone with self-supervised ImageNet-pretraining~\cite{mixmim}.

\noindent\textbf{Finetuning}\quad
We keep the pretrained encoder, abandon the decoders, and adopt state-of-the-art frameworks for finetuning.
We mainly evaluate the performance of the finetuned models on the 3D detection task on the nuScenes dataset. 
We also assess the transferability of GeoMIM on other downstream tasks.

For the 3D detection on the nuScenes dataset, we utilize the BEVDepth~\cite{li2022bevdepth} framework with an input size of $512\times1408$ for comparison with other state-of-the-art approaches. For ablation studies, we use the BEVDet~\cite{bevdet} framework with an input size of $256\times704$. 
For 3D detection on the Open Waymo dataset, the DfM~\cite{dfm,mv_fcos3d} framework is utilized. 
For 3D segmentation on the nuScenes dataset, we utilize the recent TPVFormer~\cite{tpvformer} for finetuning. We use Mask-RCNN~\cite{maskrcnn} for object detection and instance segmentation on nuImages. We follow those frameworks' default settings for finetuning, and include the detailed hyperparameters settings in the supplementary. 

We use AdamW~\cite{adamw} optimizer with a learning rate of $2 \times 10^{-4}$, weight decay of 0.05. We use layer-wise learning rate decay schedule~\cite{electra} and drop path regularization~\cite{droppath} for finetuning. We sweep the layer-wise decay rate from \{0.8, 0.85\}. The drop path rate is set to 0.2 for Swin-Base and 0.35 for Swin-Large.

\section{Main Results}

In this section, we compare our GeoMIM to prior arts on various benchmarks. We first conduct comparisons between GeoMIM and previous pretraining approaches in Sec.~\ref{sec:exp:pretrain}. We then compare our best results with state-of-the-art results on the nuScenes 3D detection benchmark in Sec.~\ref{sec:exp:sota}. To show the transferability of GeoMIM, we present the transfer learning results on other 3 benchmarks in Sec.~\ref{sec:exp:transfer}. We finally show the quantitative results in Sec.~\ref{sec:exp:visual}.

\begin{table*}
    \centering
    \resizebox{\textwidth}{!}{
    \begin{tabular}{l|c|c|c|c|c|c|c@{\hspace{1.0\tabcolsep}}c@{\hspace{1.0\tabcolsep}}c@{\hspace{1.0\tabcolsep}}c@{\hspace{1.0\tabcolsep}}c} 
    \toprule
    \textbf{Methods} & \textbf{Backbone} & \textbf{Image Size} & \textbf{Extra Data} & \textbf{TTA} & \textbf{mAP}$\uparrow$  &\textbf{NDS}$\uparrow$  & \textbf{mATE}$\downarrow$ & \textbf{mASE}$\downarrow$   &\textbf{mAOE}$\downarrow$   &\textbf{mAVE}$\downarrow$   &\textbf{mAAE}$\downarrow$  \\
    \midrule
    FCOS3D~\cite{fcos3d}         & R101-DCN   & 900 $\times$ 1600 & \xmark & \cmark & 0.358 & 0.428 & 0.690 & 0.249 & 0.452 & 1.434 & 0.124 \\
    DETR3D~\cite{detr3d}         & V2-99      & 900 $\times$ 1600 & \cmark & \cmark & 0.412 & 0.479 & 0.641 & 0.255 & 0.394 & 0.845 & 0.133 \\
    UVTR~\cite{uvtr}           & V2-99      & 900 $\times$ 1600 & \cmark & \xmark & 0.472 & 0.551 & 0.577 & 0.253 & 0.391 & 0.508 & 0.123 \\
    BEVFormer~\cite{bevformer}      & V2-99      & 900 $\times$ 1600 & \cmark & \xmark & 0.481 & 0.569 & 0.582 & 0.256 & 0.375 & 0.378 & 0.126 \\
    BEVDet4D~\cite{bevdet4d}       & Swin-B     & 900 $\times$ 1600 & \xmark & \cmark & 0.451 & 0.569 & 0.511 & 0.241 & 0.386 & 0.301 & 0.121 \\
    PolarFormer~\cite{polarformer}    & V2-99      & 900 $\times$ 1600 & \cmark & \xmark & 0.493 & 0.572 & 0.556 & 0.256 & 0.364 & 0.439 & 0.127 \\
    PETRv2~\cite{petrv2}         & GLOM-like  & 640 $\times$ 1600 & \xmark & \xmark & 0.512 & 0.592 & 0.547 & 0.242 & 0.360 & 0.367 & 0.126 \\
    BEVDepth~\cite{li2022bevdepth}       & ConvNeXt-B & 640 $\times$ 1600 & \xmark & \xmark & 0.520 & 0.609 & 0.445 & 0.243 & 0.352 & 0.347 & 0.127 \\
    BEVStereo~\cite{bevstereo}      & V2-99      & 640 $\times$ 1600 & \cmark & \xmark & 0.525 & 0.610 & 0.431 & 0.246 & 0.358 & 0.357 & 0.138 \\
    SOLOFusion~\cite{solofusion}     & ConvNeXt-B & 640 $\times$ 1600 & \xmark & \xmark & 0.540 & 0.619 & 0.453 & 0.257 & 0.376 & 0.276 & 0.148 \\
    BEVDistill~\cite{bevdistill} & ConvNeXt-B & 640 $\times$ 1600 & \xmark & \xmark & 0.498 & 0.594 & 0.472 & 0.247 & 0.378 & 0.326 & 0.125  \\
    \rowcolor[gray]{.9} 
    GeoMIM     & Swin-B & 512 $\times$ 1408 & \xmark & \xmark & 0.547 & 0.626 & 0.413 & 0.241 & 0.421 & 0.272 & 0.127 \\
    \rowcolor[gray]{.9} 
    GeoMIM     & Swin-L & 512 $\times$ 1408 & \xmark & \xmark & \textbf{0.561} & \textbf{0.644} & \textbf{0.400} & \textbf{0.238} & \textbf{0.348} & \textbf{0.255} & \textbf{0.120} \\
    \bottomrule
    \end{tabular}}
    
    \vspace{-1em}
    \caption{Comparison on nuScenes \texttt{test} set. ``Extra data" denotes depth pretraining. ``TTA" denotes test-time augmentation.}
\label{tab:main_test_set}
\end{table*}

\subsection{Comparison with previous camera-based pretraining methods}
\label{sec:exp:pretrain}

We compare our pretraining method, GeoMIM, with previous pretraining approaches to demonstrate its effectiveness in multi-view camera-based 3D detection. Four pretraining approaches for camera-based models are utilized, including the supervised pretraining on ImageNet-1K~\cite{swin}, the contrastive approach EsViT~\cite{esvit}, the multi-modal approach UniCL~\cite{unicl}, and masked-image-modeling approach MixMAE~\cite{mixmim}. Using the BEVDet framework with input size of $256\times704$, we finetune the pretrained Swin-B~\cite{swin} models on nuScenes~\cite{nuscenes} and compare their performances in Tab.~\ref{tab:pretrain}. Our approach outperforms other compared approaches in terms of all reported metrics, demonstrating the effectiveness of our pretraining method. 

Particularly, our approach achieves 0.472 NDS (NuScenes Detection Score), 2.9\% better than the self-supervised pretraining. Notably, our approach is much better at predicting translation, demonstrating a 3.3\% improvement in mATE, which shows that our geometry-enhanced pretraining can help more with localization. Surprisingly, while the contrastive or multi-modal approaches perform much better than the supervised ImageNet pretraining on various 2D visual tasks, they fail to improve the ImageNet-supervised pretraining on the 3D detection task.

\subsection{Comparison with state-of-the-art results}
\label{sec:exp:sota}

Tab.~\ref{tab:main_val_set} shows the comparison of our approach with state-of-the-art methods on nuScenes $\texttt{val}$ set. Our approach achieves state-of-the-art results of 0.605 NDS and 0.523 mAP, demonstrating substantial 2.3\% NDS and 4.0\% mAP improvements over SOLOFusion~\cite{solofusion}. Particularly, the most improvement of NDS comes from the mATE, which improves SOLOFusion by 2.7\%. 
Compared to BEVDepth~\cite{li2022bevdepth} using the same Swin-B backbone, we improve the NDS and mAP by 5.0\% and 5.7\% respectively.

On the $\texttt{test}$ set, our single model achieves 64.4\% NDS and 56.1\% mAP without using extra data and test-time augmentation, which are 2.5\% and 2.1\% better than the previous state-of-the-art results. Notably, this model performs best among all reported metrics. Compared to BEVStereo~\cite{bevstereo}, the most significant improvement of NDS comes from the mAVE (10.2\%), which
shows that our geometry-enhanced pretraining is not only better for localization but also improves the velocity estimation. 
Compared to SOLOFusion, we largely improve the mATE metric (5.3\%), showing that our pretraining is beneficial for localization. 

We also show that our pretraining is scalable in terms of model size. In particular, on the $\texttt{test}$ set, we obtain 1.8\% NDS and 1.5\% mAP gains by using the larger Swin-L~\cite{swin} backbone.

\subsection{Transfer to various 3D understanding tasks}
\label{sec:exp:transfer}

In this section, we evaluate the transferability of our approach to other datasets and tasks with different frameworks. We use three benchmarks, 3D segmentation on nuScenes dataset~\cite{nuscenes}, 3D detection on Open Waymo dataset~\cite{waymo}, and object detection and instance segmentation on nuImages dataset. 

As shown in Tab.~\ref{tab:transfer}, our approach achieves superior results on all three benchmarks, demonstrating the transferability of our pretraining method. Particularly, on the 3D segmentation task, our approach achieves 68.9\% mIoU on the nuScenes \texttt{val} set, surpassing the ImageNet-supervised pretraining~\cite{swin} results by a large margin.
Note that, unlike the 3D detection task, the self-supervised pretraining~\cite{mixmim} fails to improve the supervised pretraining because the segmentation task relies more on semantic understanding. In comparison, GeoMIM improves the ImageNet-supervised pretraining for 2.5\% mIoU. On the nuScenes \texttt{test} test, we achieve state-of-the-art results, 1.1\% mIoU better than the previous best camera-based results in TPVFormer~\cite{tpvformer}.
Moreover, our pretrained backbone can also transfer to datasets that differ from that used in pretraining. On Open Waymo detection benchmark, our pretraining improves the MixMAE~\cite{mixmim} self-supervised pretrained model by 3.0\%/3.0\% on LET-3D APL/AP. 

Apart from the 3D perception task, we show that our pretrain can also transfer to 2D object detection and instance segmentation tasks. As shown in Tab.~\ref{tab:transfer} (right), GeoMIM improves the self-supervised pretraining by 1.4\% AP\textsuperscript{box} and 2.6\% AP\textsuperscript{mask}.

\section{Ablation Studies}

In this section, we conduct ablation studies to evaluate the impact of different design choices on the performance of our proposed GeoMIM on the multi-view camera-based 3D detection task. Unless otherwise specified, we use the Swin-B~\cite{swin} backbone and pretrain it for 6 epochs. We utilize the BEVDet~\cite{bevdet} framework for finetuning the pretrained backbone and report the performance on the nuScenes \texttt{val} set~\cite{nuscenes}. The gray column indicates the final choice of GeoMIM.

\noindent\textbf{Pretraining epochs and pretraining data.}\quad  We explore the effect of pretraining epochs and pretraining data on GeoMIM. As shown in Tab.~\ref{tab:aba:epoch}, we find that we can improve the mATE performance but degenerate the mAOE performance through 6 epochs of pretraining. Interestingly, if we pretrain for more epochs, mATE performance saturates but mAOE can be largely improved. Additionally, as shown in Tab.~\ref{tab:aba:data}, the performance of all metrics gradually increases as we use more data for pretraining.

We also compare the data utilization ability of different pretraining in Tab.~\ref{tab:aba:data_utilization}. If we use self-supervised pretraining MixMAE~\cite{mixmim}, the performance gain between using 50\% and 100\% data is 4.2\% NDS. If we use GeoMIM pretraining, the improvement increases to 5.1\% NDS. The results demonstrate that our GeoMIM can be benefited more by using more data.

\begin{table}[t]
    \centering
    \resizebox{0.8\linewidth}{!}{
    \begin{tabular}{l|c|c|cc} 
    \toprule
    \textbf{\# epochs} & \textbf{NDS}$\uparrow$ & \textbf{mAP}$\uparrow$ & \textbf{mATE}$\downarrow$ & \textbf{mAOE}$\downarrow$ \\
    \midrule
    0 & 0.443 & 0.374 & 0.647 & 0.418 \\
    6 & 0.460 & 0.381 & \textbf{0.613} & 0.449 \\
    \rowcolor[gray]{.9} 
    50 & 0.472 & 0.398 & 0.614 & 0.395 \\
    100 & \textbf{0.475} & \textbf{0.400} & 0.615 & \textbf{0.390} \\
    \bottomrule
    \end{tabular}}
    \vspace{-1em}
    \caption{Ablation of petraining epochs.}
    \label{tab:aba:epoch}
\end{table}

\begin{table}[t]
    \centering
    \resizebox{0.8\linewidth}{!}{
    \begin{tabular}{l|c|c|cc} 
    \toprule
    \textbf{\%data} & \textbf{NDS}$\uparrow$ & \textbf{mAP}$\uparrow$ & \textbf{mATE}$\downarrow$ & \textbf{mAOE}$\downarrow$ \\
    \midrule
    0 & 0.443 & 0.374 & 0.647 & \textbf{0.418} \\
    10 & 0.426 & 0.356 & 0.646 & 0.519 \\
    50 & 0.445 & 0.367 & 0.623 & 0.465 \\
    \rowcolor[gray]{.9} 
    100 & \textbf{0.460} & \textbf{0.381} & \textbf{0.613} & 0.449 \\
    \bottomrule
    \end{tabular}}
    \vspace{-1em}
    \caption{Ablation of the percentage of the pretraining data.}
    \label{tab:aba:data}
\end{table}

\begin{table}[t]
    \centering
    \resizebox{0.7\linewidth}{!}{
    \begin{tabular}{cc|cc} 
    \toprule
    \textbf{\% data} & \textbf{w/ GeoMIM} & \textbf{NDS}$\uparrow$ & \textbf{mAP}$\uparrow$ \\
    \midrule
    50\% & \multirow{2}{*}{\xmark} & 0.401 & 0.355 \\
    100\% &  & 0.443 & 0.374 \\
    \midrule
    50\% & \multirow{2}{*}{\cmark} & 0.409 & 0.356 \\
    100\% &  & 0.460 & 0.381 \\
    \bottomrule
    \end{tabular}
    }
    \vspace{-1em}
    \caption{Ablation of data utilization.}
    \label{tab:aba:data_utilization}
\end{table}

\begin{table}[t]
    \centering
    \resizebox{0.8\linewidth}{!}{
    \begin{tabular}{l|c|c|cc} 
    \toprule
    \textbf{Mask ratio} & \textbf{NDS}$\uparrow$ & \textbf{mAP}$\uparrow$ & \textbf{mATE}$\downarrow$ & \textbf{mAOE}$\downarrow$ \\
    \midrule
    0.25  & 0.454 & 0.375 & 0.622 & \textbf{0.447} \\
    \rowcolor[gray]{.9} 
    0.5 & \textbf{0.460} & \textbf{0.381} & \textbf{0.613} & 0.449 \\
    0.75 & 0.453 & 0.375 & 0.642 & 0.452 \\
    \bottomrule
    \end{tabular}}
    \vspace{-1em}
    \caption{Ablation of the mask ratio.}
    \label{tab:aba:mask}
\end{table}

\noindent\textbf{Mask ratio.}\quad
We examine the effect of the mask ratio used in the masked image modeling training process on the performance. As shown in Tab.~\ref{tab:aba:mask}, we find that using a mask ratio of 0.5 performs best as a very high mask ratio causes the pretext task too hard while a low mask ratio causes the pretext task too easy.

\noindent\textbf{Pretraining with distillation or other reconstruction targets.}\quad
We compare the performance of GeoMIM with different learning targets, including RGB pixels~\cite{mae} and the voxelized LiDAR points. Moreover, we use the depth ground truth derived from the LiDAR point cloud for depth pretraining~\cite{dd3d}. Following previous works, we also the LiDAR BEV features for conducting distillation pretraining~\cite{liga,bevdistill,cmkd}. 
We initialize the backbone with MixMAE~\cite{mixmim} self-supervised pretraining and use its results for comparison. All the pretraining experiments are conducted on the nuScenes dataset. 

As shown in Tab.~\ref{tab:aba:target}, we find the depth or distillation pretraining fails to improve the MixMAE results. Those two pretraining methods are beneficial for object localization to improve the mATE metric, but degenerate mAOE a lot. Using the RGB pixels as the reconstruction targets like MAE is also unable to improve the NDS. The main reason is that the nuScenes dataset is much less diverse than the widely used ImageNet-1K~\cite{imagenet} dataset, and as a result, the model is easy to overfit the training data. Moreover, though the LiDAR points contain rich geometry information, we find that directly using the voxelized LiDAR points as the reconstruction targets also fails to improve the MixMAE results. As stated in Sec.~\ref{sec:intro}, the LiDAR voxels are sparse and noisy. Using them as the reconstruction targets results in unstable pretraining. In comparison, we use a pretrained LiDAR detection model to extract more meaningful BEV features as the reconstruction targets, which can not only transfer the rich geometry information to the camera-based model but also avoid the noise problem of directly reconstructing LiDAR voxels. 

\begin{table}[t]
    \resizebox{1.0\linewidth}{!}{
    \begin{tabular}{l|c|c|cc} 
    \toprule
    \textbf{Approach} & \textbf{NDS}$\uparrow$ & \textbf{mAP}$\uparrow$ & \textbf{mATE}$\downarrow$ & \textbf{mAOE}$\downarrow$ \\
    \midrule
    SSL Pretrain~\cite{mixmim} & 0.443 & 0.374 & 0.647 & \textbf{0.418} \\
    \midrule
    + Depth Pretrain & 0.426 & 0.350 & 0.623 & 0.535 \\
    + Distill Pretrain & 0.434 & 0.360 & 0.626 & 0.533 \\
    + LiDAR reconstruction & 0.439 & 0.352 & 0.637 & 0.483 \\
    + RGB reconstruction & 0.440 & 0.376 & 0.632 & 0.484 \\
    \rowcolor[gray]{.9} 
    + GeoMIM & \textbf{0.460} & \textbf{0.381} & \textbf{0.613} & 0.449 \\
    \bottomrule
    \end{tabular}}
    \vspace{-1em}
    \caption{Ablation of pretraining setups and targets.}
    \label{tab:aba:target}
\end{table}

\begin{table}[t]
    \centering
    \resizebox{0.9\linewidth}{!}{
    \begin{tabular}{ccc|cc|cc} 
    \toprule
    \multirow{2}{*}{\textbf{Decouple}} & \multirow{2}{*}{\textbf{CVA}} & \multirow{2}{*}{\textbf{Cam.}} & \multicolumn{2}{c|}{\textbf{nuScenes}} & \multicolumn{2}{c}{\textbf{Waymo}} \\
     & &  & \textbf{NDS} & \textbf{mAP} & \textbf{APL} & \textbf{AP} \\
    \midrule
    \rowcolor[gray]{.9} 
    \cmark & \cmark & \cmark & \textbf{46.0} & \textbf{38.1} & \textbf{37.0} & \textbf{51.6} \\
    \textcolor{gray}{\xmark} & \cmark & \cmark & 45.2 & 37.8 & 36.4 & 51.4 \\
    \cmark & \textcolor{gray}{\xmark} & \cmark & 45.5 & \textbf{38.1} & 36.8 & 51.3 \\
    \cmark & \textcolor{gray}{\xmark} & \textcolor{gray}{\xmark} & 45.1 & 37.9 & 36.0 & 51.0 \\
    \bottomrule
    \end{tabular}}
    \vspace{-1em}
    \caption{Ablation of the decoder designs. ``Cam." denotes the camera-aware design.}
    \label{tab:aba:decoder}
\end{table}

\begin{table}[t]
    \resizebox{1.0\linewidth}{!}{
    \begin{tabular}{l|c|cccc} 
    \toprule
    \textbf{Backbone} & \textbf{Parameters} & \textbf{NDS}$\uparrow$ & \textbf{mAP}$\uparrow$ & \textbf{mATE}$\downarrow$ & \textbf{mAOE}$\downarrow$ \\
    \midrule
    Swin-B & 88M & 0.460 & 0.381 & 0.613 & 0.449 \\
    Swin-L & 197M & \textbf{0.478} & \textbf{0.398} & \textbf{0.609} & \textbf{0.371} \\
    \bottomrule
    \end{tabular}}
    \vspace{-1em}
    \caption{Ablation of the model sizes.}
    \label{tab:aba:backbone}
\end{table}

\begin{figure*}
    \centering
    \includegraphics[width=1.0\linewidth]{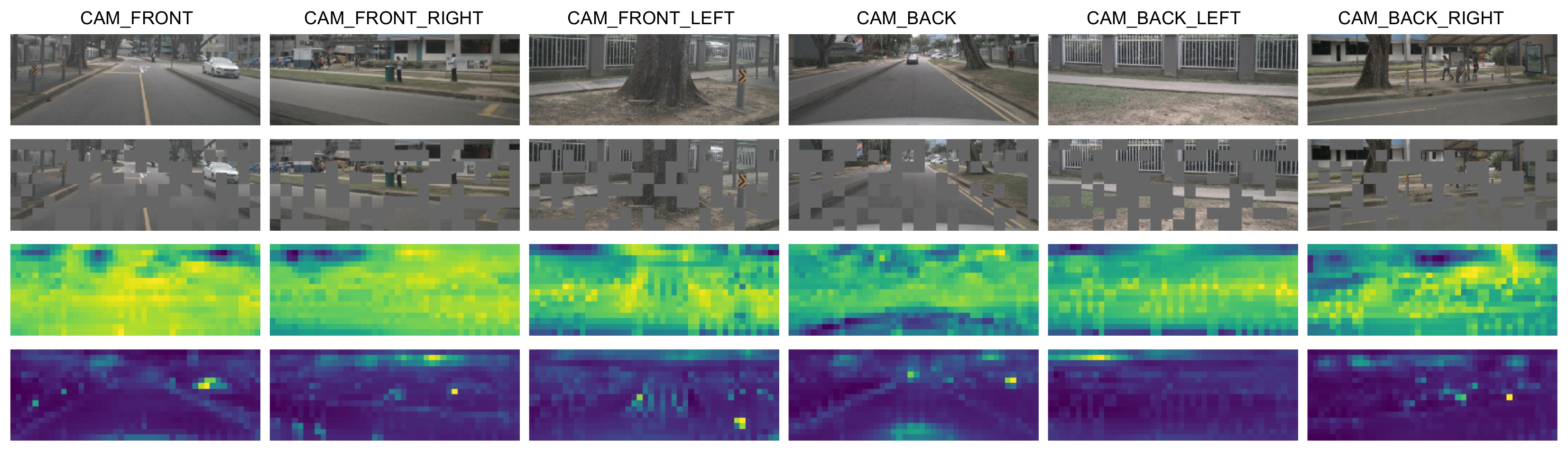}
    \vspace{-1em}
    \caption{Example results on nuScenes \texttt{val} images. From top to bottom, the rows are the camera-view image, masked camera-view image, decoded semantic features, and decoded geometry features. }
    \label{fig:rec}
\end{figure*}

\begin{figure*}
    \centering
    \includegraphics[width=1.0\linewidth]{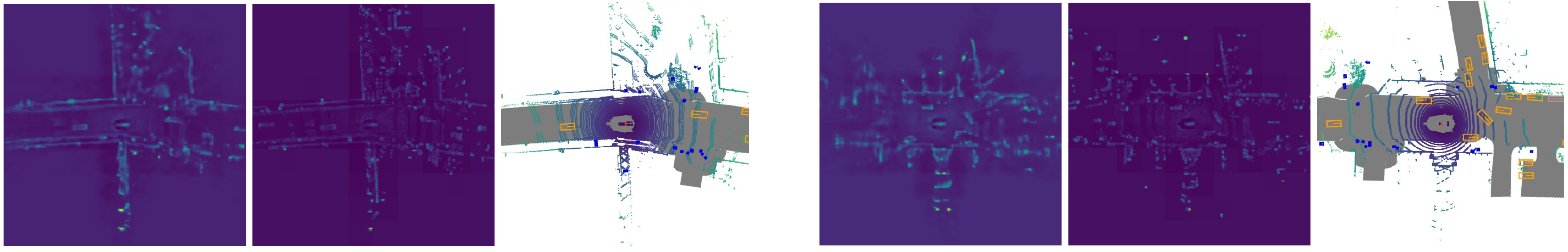}
    \vspace{-1em}
    \caption{Reconstruction results on nuScenes \texttt{val} images. For each triplet, we show the reconstructed BEV features (left), LiDAR BEV features, and LiDAR point cloud (right) in BEV.}
    \label{fig:bev_rec}
\end{figure*}

\begin{figure}
    \centering
    \includegraphics[width=1.0\linewidth]{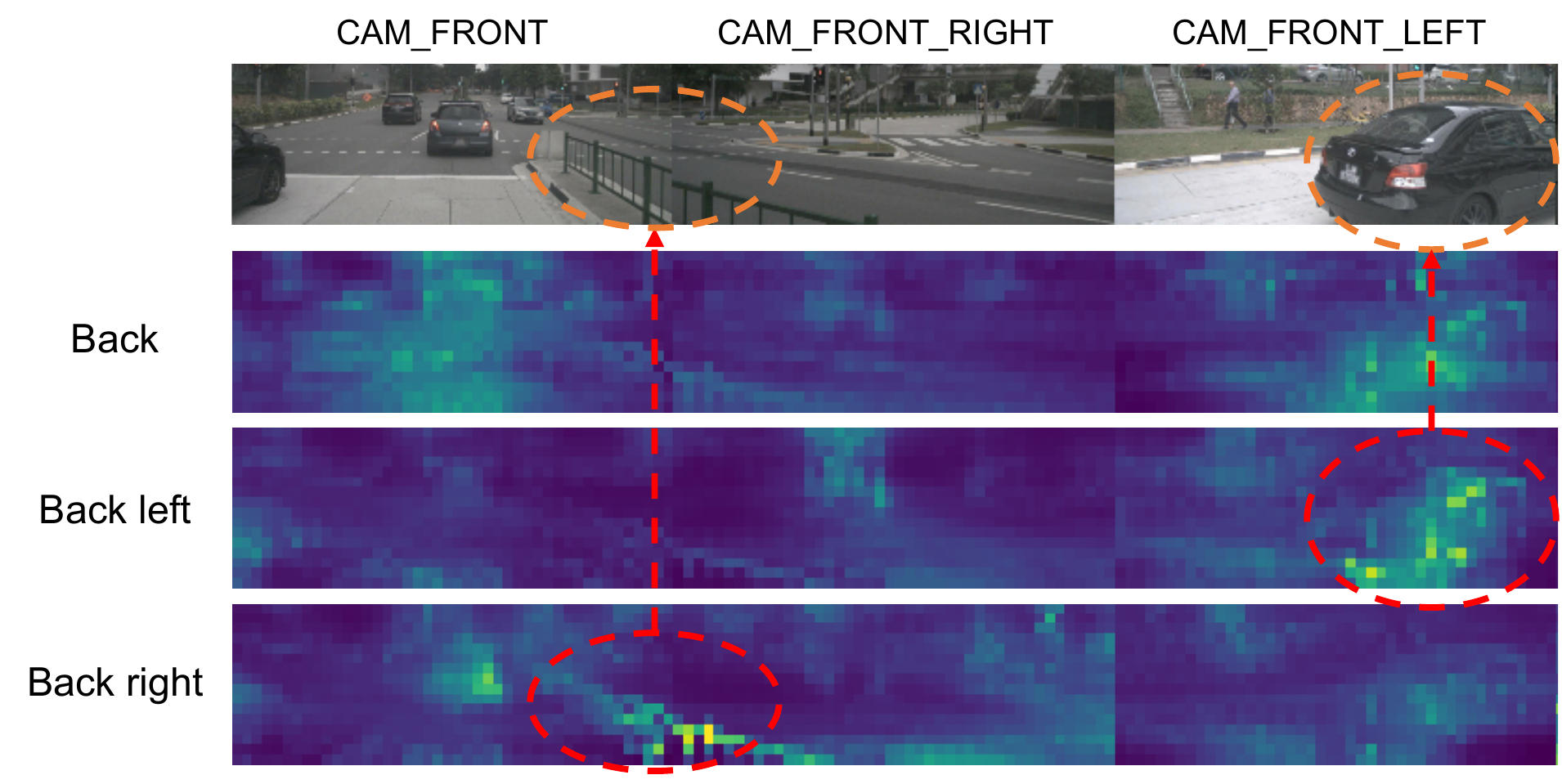}
    \vspace{-1em}
    \caption{CVA blocks' attention maps across different cameras. 
    Our CVA enables the cameras at back to attend to the semantic parts in front views.}
    \label{fig:cva_map}
\end{figure}

\noindent\textbf{Ablation on the decode designs.}\quad
We investigate the impact of the proposed decoder designs on the final performance. Apart from reporting results on the nuScenes dataset, we also report the performance on the Open Waymo dataset to show how the design choices affect the transferability of the pretraining. 

As shown in Tab.~\ref{tab:aba:decoder}, we find that using a decoupled decoder to separately decode the depth and semantic features can improve the NDS for 0.8\%, compared to using one decoder to jointly decode them. The decoupled branches force the geometry branch to focus on inferring depth from geometry cues, which maximizes the utilization of the model capacity. Moreover, removing the Cross-View Attention (CVA) blocks results in a 0.5\% performance drop because of lacking cross-view interaction for better BEV feature inference. Additionally, we find that further removing the camera-aware design leads to 0.8\% LET-3D APL drop on the Open Waymo dataset. As the pretraining and finetuning might be conducted on different datasets, using camera-aware depth reconstruction is beneficial for the transferability of the pretraining. 

We also experiment to consider epipolar geometry in our proposed CVA block. We observed a slight accuracy improvement (0.463 NDS vs. 0.460 NDS). However, including epipolar geometry constraints in CVA block slowed down the training by
1.5 times due to the slower sampling along the epipolar plane on GPUs.

\noindent\textbf{Ablation on backbone size.}\quad We investigate the scalability of our pretraining method and show the results in Tab.~\ref{tab:aba:backbone}. 
Our GeoMIM is capable to be scaled up to Swin-L model with 200M parameters. The pretrained Swin-L~\cite{swin} improves Swin-B on all reported metrics, especially the mAOE performance (7.8\%).

\section{Qualitative Evaluation}
\label{sec:exp:visual}
\noindent\textbf{Reconstruction visualization.}\quad We show the decoded geometry features and semantic features during pretraining in Fig.~\ref{fig:rec}. Although a high mask ratio is utilized, meaningful patterns can still be observed in the decoded features. In particular, we find the geometry branch focuses on the structure of the driving scenes while the semantic branch can be activated by different semantic regions, including roads, cars, trees, etc. Furthermore, we visualize the reconstructed BEV features in Fig.~\ref{fig:bev_rec}. The reconstructed BEV features can well restore the LiDAR model's BEV features, including the road structures and the semantic features.

\noindent\textbf{Cross-view attention maps.}\quad 
We visualize the attention maps of the Cross-View Attention (CVA) blocks in Fig.~\ref{fig:cva_map}. Through the cross-view interaction, one view is able to attend to the semantic parts of other views. 

\begin{figure}
    \centering
    \includegraphics[width=0.7\linewidth]{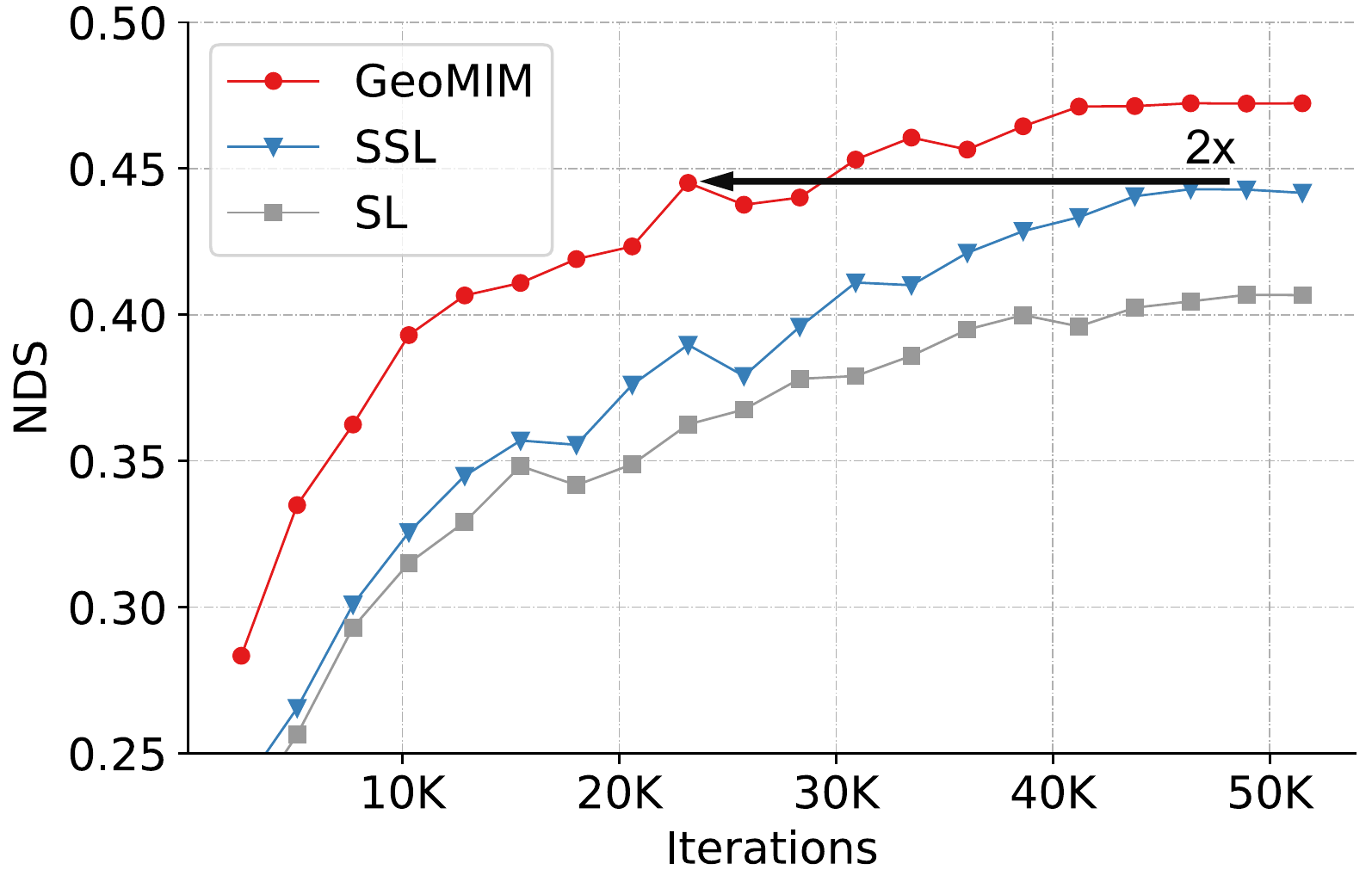}
    \vspace{-1em}
    \caption{Performance curves of different pretraining methods. ``SL" and ``SSL" denote the ImageNet-supervised and MixMAE self-supervised pretraining respectively.}
    \vspace{-1em}
    \label{fig:curve}
\end{figure}

\noindent\textbf{Convergence curve.}\quad  We show the convergence curve of different pretraining in Fig.~\ref{fig:curve}. Our pretraining can largely improve the self-supervised~\cite{mixmim} and supervised~\cite{swin} pretraining in terms of the convergence speed, which can match the self-supervised pretraining's final results with only half of the iterations.

\section{Conclusion}
In this paper, we proposed a pretraining method, GeoMIM, for multi-view camera-based 3D detection. By leveraging the knowledge of a pretrained LiDAR model in a pretrain-finetune paradigm, GeoMIM aims to transfer its rich geometry knowledge to the camera-based model. Specifically, GeoMIM reconstructs BEV features from masked images via a novel decoder and a cross-view attention mechanism. We demonstrate that GeoMIM significantly outperforms existing state-of-the-art methods on the nuScenes dataset, achieving state-of-the-art results in both camera-based 3D detection and segmentation tasks. Moreover, we verify that the pretrained model can be transferred to the Waymo Open dataset, further showing its effectiveness and generality.

\noindent\textbf{Limitations}\quad
Despite the promising results, GeoMIM also has some limitations. First, GeoMIM requires a large amount of labeled data for pretraining, which may not be available in some applications. Second, GeoMIM relies on the quality of the LiDAR model's BEV features, which may not always be accurate or complete. Overall, while GeoMIM shows great potential, further research is needed to address these limitations and improve its applicability in a wider range of applications.

\paragraph{Acknowledgement}
This project is funded in part by National Key R\&D Program of China Project 2022ZD0161100, by the Centre for Perceptual and Interactive Intelligence (CPII) Ltd under the Innovation and Technology Commission (ITC)’s InnoHK, by General Research Fund of Hong Kong RGC Project 14204021. Hongsheng Li is a PI of CPII under the InnoHK.

{\small
\bibliographystyle{ieee_fullname}
\bibliography{egbib}
}

\end{document}